\documentclass[twoside,11pt]{article}

\usepackage{jmlr2e}
\usepackage{fancyvrb}
\newcommand{\pkg}[1]{{\fontseries{b}\selectfont #1}}

\ShortHeadings{Comparing Structure Learning Algorithms}{Ramsey, Malinsky, Bui}
\firstpageno{1}

\begin{document}

\title{\pkg{algcomparison}: Comparing the Performance of Graphical Structure Learning Algorithms with TETRAD}

\author{\name Joseph D. Ramsey \email jdramsey@andrew.cmu.edu \\
       \addr Department of Philosophy\\
       Carnegie Mellon University\\
       Pittsburgh, PA USA
       \AND
       \name Daniel Malinsky \email dsm2128@cumc.columbia.edu \\
       \addr Department of Biostatistics\\
       Columbia University\\
       New York, NY USA
       \AND
   	   \name Kevin V. Bui \email kvb2@pitt.edu \\
   	   \addr Department of Biomedical Informatics\\
   	   University of Pittsburgh\\
   	   Pittsburgh, PA USA}

\editor{editor}

\maketitle

\begin{abstract}%   <- trailing '%' for backward compatibility of .sty file
In this report we describe a tool for comparing the performance of graphical causal structure learning algorithms implemented in the TETRAD freeware suite of causal analysis methods. Currently the tool is available as package in the TETRAD source code (written in {Java}). Simulations can be done varying the number of runs, sample sizes, and data modalities. Performance on this simulated data can then be compared for a number of algorithms, with parameters varied and with performance statistics as selected, producing a publishable report. 
%The order of the algorithms in the output can be adjusted to the user's preference using a utility function over the statistics. Data sets from simulation can be saved along with their graphs to a file and loaded back in for further analysis, or used for analysis by other tools. 
The package presented here may also be used to compare structure learning methods across platforms and programming languages, i.e., to compare algorithms implemented in TETRAD with those implemented in {MATLAB}, {Python}, or {R}.
\end{abstract}


\section{Introduction}

Often researchers are faced with the problem of choosing an algorithm from among
possibly dozens of relevant algorithms for a particular task. This can be
time-consuming and error-prone; one must try each algorithm in turn, vary the
parameters for that algorithm, run it in simulation on common data sets that
hopefully reflect the properties of the real data of interest, and
discern which algorithm has the best performance over the range of cases under
study. Research papers tend to compare only a small number of algorithms
at a time on performance statistics that may not be of interest to the user,
using simulation settings not appropriate for the domain. Ideally a user could
directly compare a range of algorithms on data of their choosing and on
performance statistics of interest to them, so that they could make an informed
decision as to which algorithm(s) may be best suited to the user's particular
purpose.

We focus on the causal structure learning algorithms in the TETRAD
freeware.\footnote{\url{https://github.com/cmu-phil/tetrad}} Within TETRAD, we
have created a tool for comparing algorithms, both ``basic" algorithms with
varying parameter settings and algorithms variously combined. The relevant code
is contained in the package \pkg{algcomparison} within TETRAD.\footnote{The full
package path in the code library is \url{edu.cmu.tetrad.algcomparison}.} It is
possible to construct studies in which combinations of structure learning
algorithms are compared head-to-head on common data, with known true models;
winners conveniently float to the top of the list of algorithms when sorted by a
utility function that reflects the user's interests. Algorithms that perform
poorly for the intended type of data analysis quickly become apparent. This
makes it easy to identify the general class of algorithms the user may want to
select from for their purposes. %As the user learns more about their particular problem, they can easily refine their assessment by running a modified set of simulations. 
%The procedure described here is more time-efficient than setting up specific simulation tests for pairs or triples of algorithms, since one can compare simultaneously as many algorithms and algorithm variants as one wishes.
%A general problem which arises in developing novel algorithms is knowing how their results compare to those of existing algorithms, where the algorithms' weaknesses are from the point of view of performance, where they need to be improved, and whether a development effort should be abandoned in favor of other methods with better performance. Having a tool to compare automatically a new algorithm to all existing algorithms, easily, on-the-fly, is of non-trivial advantage for future algorithm development.

In TETRAD, \pkg{algcomparison} has available a wide range of algorithms and the flexbility to add new algorithms easily. Combinations of existing algorithms are often treated in practice as novel algorithms; we allow them to be treated as such. \pkg{algcomparison} has some standard styles of simulation readily available and the user is able to add new simulation styles.
The tool enables a user to use ``default'' parameters and to change the default settings of the parameters easily. \pkg{algcomparison} has a range of built-in standard performance statistics for quantifying the accuracy of a learned structure, and with some straightforward programming the user may add new performance statistics. 
Finally, in deference to the user's needs, \pkg{algcomparison} enables the user to decide which combination of performance statistics to employ to determine the best algorithm or algorithms. This is because different users with different scientific backgrounds may very well have different views as to what is important in an estimated causal model.
We take the view that these differences should be handled using a modular architecture. 
Algorithms, simulations, parameters, performance statistics, independence tests, and scores can be independently input into a central comparison class to execute experiments as the user wishes.

\section{Background}

The TETRAD software was introduced in the mid-1980s to aid in constructing, testing, predicting with, and learning causal statistical models based on structural equations or graphical representations like Directed Acyclic Graphs (DAGs) \citep{glymour1987discovering, spirtes1990simulation, scheines1998tetrad, spirtes2000causation}. The capabilities and flexibility of TETRAD has increased with years of algorithm development and application in several scientific fields including biology \citep[e.g.][]{shipley2006fundamental}, neuroscience \citep[e.g.][]{smith2011network, mills2014brain}, economics \citep[e.g.][]{bessler2002money, demiralp2003searching}, climate science \citep[e.g.][]{ebert2012causal}, education research \citep[e.g.][]{rau2013does}, and other areas. Though TETRAD is capable of performing a wide range of tasks relevent to causal inference, we will focus only on graphical structure learning here.

TETRAD implements numerous algorithms which search for causal graphical models. The resultant models are intended to have a causal interpretations, the precise details of which depend on the underlying assumptions and the type of output graph produced by the method. At the time of writing, there are dozens of structure learning algorithms and variations; we do not review them all in detail here. Recent overviews of causal graphical modeling, structure learning algorithms, and the assumptions underlying causal inference from observational data can be found in \cite{spirtes2016causal}, \cite{drton2017structure}, and \cite{heinze2018causal}. Popular alternative software packages for structure learning include \pkg{pcalg} \citep{kalisch2012causal} and \pkg{bnlearn} \citep{scutari2010} in R and the \pkg{Bayes Net Toolbox} in MATLAB \citep{murphy2001bayes}. These packages implement some of the same algorithms available in TETRAD, as well as some methods which are not currently available in TETRAD. TETRAD also implements many algorithms not available on any of these alternative platforms. Although cross-evaluation of multiple algorithms is possible using (for example) \pkg{pcalg} as a base,\footnote{\url{https://github.com/christinaheinze/CompareCausalNetworks}} the benefit of \pkg{algcomparison} is its flexibility and the fact that basically no novel programming is required on behalf of the user; the tool can be executed entirely on the command line with simple XML scripts as we describe below. 

\section{A modular architecture}

The source code is structured around several Java interfaces, specifically interfaces for specifying search algorithms (\texttt{Algorithm}), simulations and their parameters (\texttt{Simulation}), parametric or nonparametric conditional independence tests (\texttt{IndependenceWrapper}), model scores (\texttt{ScoreWrapper}), and so on, along with a special class called \texttt{Comparison} which contains methods to load in files or data and execute a sequence of simulations, algorithms, and comparisons for given parameter settings. A \texttt{Simulation} method generates a random graph (e.g., by adding edges between vertices arranged in some partial order with fixed probability) and then generates samples according to that graph by some scheme (e.g., a linear or nonlinear structural equation model, or a multinomial model for discrete data). An \texttt{Algorithm} is any method which takes data and parameter settings as input and returns a graphical representation: these may be constraint-based methods which use independence tests (e.g., PC, FCI and their variants), score-based methods which optimize some model score (e.g., Fast Greedy Equivalence Search and related), pairwise methods based on regression residual asymmetries, some novel user-specified algorithm, or a combination of these. The software also includes some methods for learning undirected graphs with no obvious causal interpretation (e.g., graphical Lasso), since these may be useful as subroutines of a causal learning procedure.%The user may also define new instances of any of the aforementioned interfaces to evaluate novel proposals or techniques.

\section{Running comparison experiments}

There are two ways to run comparisons with \pkg{algcomparison}. The first is by executing a short Java script within TETRAD, typically in an integrated development environment (IDE). Several example scripts are bundled with the code and may be modified by the user. The second is by running an XML configuration file on the command line, which does not require any knowledge of Java programming.\footnote{The code for the command line interface is separate from TETRAD and can be found here:\\ \url{https://github.com/bd2kccd/causal-compare}.} In both cases, the modular components described above are combined according to the user's specifications and passed to the \texttt{Comparison} class. An example XML configuration file is as follows:
\begin{small}
\begin{Verbatim}
<comparison> <compareBy>
	<search>
		<simulations>
			<simulation source="generate">
			<graphtype>RandomForward</graphtype>
			<modeltype>SemSimulation</modeltype>
			</simulation>
		</simulations>
		<algorithms>
			<algorithm name="pc-all">
			<test>fisher-z-test</test>
			<score>sem-bic</score>
			</algorithm>
			<algorithm name="fges">
			<score>sem-bic</score>
			</algorithm>
		</algorithms>
		<parameters>
		<parameter name="numMeasures">1000</parameter>
		<parameter name="avgDegree">4</parameter>
		</parameters>
	</search> </compareBy>
	<statistics>
		<statistic>adjacencyPrecision</statistic>
		<statistic>adjacencyRecall</statistic>
		<statistic>SHD</statistic>
	</statistics>
</comparison>
\end{Verbatim}
\end{small}
%Describe the main interfaces (algorithm, simulation, indtest, scores, etc), and how they work together.
%<properties>
%<property name="showAlgorithmIndices">true</property>
%</properties>
This generates data from a linear Gaussian model with 1000 variables and compares the PC algorithm with Fisher Z independence test to the FGES algorithm, which uses the BIC score. It calculates adjacnecy precision, recall, and Strutural Hamming Distance from the true model (its Markov equivalence class). Running this configuration on the command line would generate a table of results 
%(with averages, standard deviations, and worst-case analysis over runs) 
saved to an output file. There is a wide range of algorithms, settings, tests, and statistics available within TETRAD which we do not have space to enumerate here, though a user may 
%consult the various example scripts provided with the source code or 
generate a list of available options by executing the script called \texttt{RunConfig}.

\section{Cross-platform comparisons}

It may be desirable to
assess algorithms implemented in alternative languages and software platforms. Furthermore, one may wish to evaluate how
TETRAD algorithms perform on data generated using other software packages.
%While it is not always possible to drive algorithms on other platforms from Java, nonetheless 
\pkg{algcomparison} will save data and graphs to disk in a common
format. These data and graphs can be loaded and analyzed in
other platforms, such as R, MATLAB, or Python. The results of these analyses can
then be stored, read by \pkg{algcomparison}, and
included in comparison tables alongside TETRAD results. Data
simulated on other platforms may also be loaded into the \pkg{algcomparison}
tool. Instructions for executing cross-platform comparison are found on the project page: \url{https://github.com/bd2kccd/causal-compare}. 

\newpage

% Acknowledgements should go at the end, before appendices and references

\acks{Research reported in this publication was supported by grant U54HG008540 awarded by the National Human Genome Research Institute through funds provided by the trans-NIH Big Data to Knowledge (BD2K) initiative.}


%\newpage

%\appendix
%\section*{Appendix A.}
%\label{app:theorem}

% Note: in this sample, the section number is hard-coded in. Following
% proper LaTeX conventions, it should properly be coded as a reference:

%In this appendix we prove the following theorem from
%Section~\ref{sec:textree-generalization}:

%In this appendix we prove the following theorem from
%Section~6.2:

\vskip 0.2in
\bibliography{../comparison_report}

\end{document}